\pdfoutput=1

\documentclass[11pt]{article}

\usepackage{emnlp2021}

\usepackage{times}
\usepackage{latexsym}

\usepackage{booktabs}
\usepackage{tabularx}
\usepackage{hyperref}
\usepackage{xfrac}
\usepackage{dirtytalk}
\usepackage{xcolor}
\usepackage{xspace} 
\usepackage{graphicx}
\usepackage{times}
\usepackage{comment}
\usepackage{amsmath}
\usepackage{ulem}
\usepackage{enumitem}
\usepackage{makecell}
\usepackage{multirow}
\usepackage{comment}
\usepackage{amssymb}
\usepackage{float}
\usepackage{caption}
\usepackage{subcaption}

\usepackage[T1]{fontenc}

\usepackage[utf8]{inputenc}

\usepackage{microtype}

\newcommand{\noedit}[1]{``no edit#1''}
\newcommand{\neededit}[1]{``need edit#1''}
\newcommand{\token}[1]{\texttt{#1}}

\newcommand{\MATH}{\_MATH\_}
\newcommand{\spelling}{\textit{spelling error}}
\newcommand{\deleted}{\textit{deleted text}}
\newcommand{\nop}[1]{}
\newcommand{\gradientxinput}[1]{gradient$\times$input#1}
\newcommand{\gradientxinputmagnitude}[1]{$\lvert \text{gradient}\times \text{input}\rvert$#1}

\setlist[enumerate,itemize]{nosep}

\title{An Investigation of Language Model Interpretability via Sentence Editing}

\author{Samuel Stevens \\
  The Ohio State University \\
  \texttt{stevens.994@osu.edu} \\\And
  Yu Su \\
  The Ohio State University \\
  \texttt{su.809@osu.edu} \\}

\begin{document}
\maketitle
\begin{abstract}
Pre-trained language models (PLMs) like BERT are being used for almost all language-related tasks, but interpreting their behavior still remains a significant challenge and many important questions remain largely unanswered. 
In this work, we re-purpose a sentence editing dataset, where faithful high-quality human rationales can be automatically extracted and compared with extracted model rationales, as a new testbed for interpretability. This enables us to conduct a systematic investigation on an array of questions regarding PLMs' interpretability, including the role of pre-training procedure, comparison of rationale extraction methods, and different layers in the PLM.  The investigation generates new insights, for example, contrary to the common understanding, we find that attention weights correlate well with human rationales and work better than gradient-based saliency in extracting model rationales. Both the dataset and code are available at \url{https://github.com/samuelstevens/sentence-editing-interpretability} to facilitate future interpretability research.
\end{abstract}

\section{Introduction}

Pre-trained language models (PLMs)~\citep{devlin-etal-2019-bert,liu2019roberta,beltagy-etal-2019-scibert} are pervasively used in language-related tasks, but interpreting their predictions is notoriously difficult because of their parameters' complex inter-dependencies.
Given a specific prediction, we want to know \textit{why} a model made that decision, both to further improve performance and to use the model in high-stakes scenarios such as healthcare or bank loan approvals where interpretability is important.
This has motivated efforts in extracting model explanations, typically in the form of \textit{rationales}, i.e., subsets of the original input that support a decision \citep{zaidan-etal-2007-using}.
Attention heatmaps \citep{xu-etal-2015-show} and gradient-based saliency maps \citep{simonyan2014deep} are common extraction methods.

\begin{figure}[t]
  \centering
  \small
  \begin{tabularx}{\linewidth}{X}
  \toprule
    \texttt{\nop{<sentence sid="8447.0">}The algorithm <del>descripted</del> <ins>described</ins> in the previous sections has several advantages.\nop{</sentence>}} \\ 
    \midrule
    The algorithm \sout{descripted}$\to$\textbf{described} in the previous sections has several advantages.\\ 
    \midrule
    \midrule
    \texttt{\nop{<sentence sid="662.3">}However, we <del>must note that we </del>still have no means of deciding which documents out of \MATH{} deserve to be in \MATH{} and \MATH{}, respectively.\nop{</sentence>}} \\ 
    \midrule
    However, we \sout{must note that we }still have no means of deciding which documents out of \MATH{} deserve to be in \MATH{} and \MATH{}, respectively. \\
  \bottomrule
  \end{tabularx}
  \caption{
    Two \neededit{} examples from AESW in the original data format and a human-readable format.
    The first example (a) has a spelling error ``descripted'' and the second (b) is edited for concision.
  }
  \label{fig:aesw-example}
  \vspace{-10pt}
\end{figure}

There have been efforts on developing datasets for interpretability research, for example, the recent ERASER benchmark~\cite{deyoung-etal-2020-eraser}. 
However, the majority of ERASER tasks use human rationales highlighted by a different annotator after the original labeling process.
Such rationales are not necessarily \textit{faithful}; a rationale highlighted by the second annotator may not have been actually used by the first annotator while labeling.
Manual rationale labeling is also difficult and time-consuming; of the six datasets in the ERASER benchmark, only one has more than 200 examples.

Our first contribution is the realization that AESW (Automatic Evaluation of Scientific Writing; \citealp{aesw}), a sentence editing dataset, contains \textit{thousands} of \textit{faithful human rationales} that can be automatically re-purposed for interpretability research. 
See Figure \ref{fig:aesw-example} for examples.
This provides a new, large-scale dataset with truly faithful human rationales for interpretability questions surrounding model rationales.

Our second contribution is investigating multiple factors in PLM rationale plausibility.
More plausible rationales are valuable in human-in-the-loop systems where humans use model rationales to make a final decision.
We compare (1) pre-training procedures, (2) attention weight- and input gradient-based methods of extracting model rationales, (3) correlation between model rationale plausibility and model confidence, and (4) differences in transformer layers.
While previous work \citep{jain-wallace-2019-attention,serrano-smith-2019-attention} has shown that attention weights are not always faithful, we find that they correlate with human rationales better than gradient-based methods.

\section{Related Work}

Human rationales (as defined by \citealp{zaidan-etal-2007-using}) are subsets of input highlighted by human annotators as evidence to support a decision.
The same annotator labeling an example might also highlight their rationale \citep{khashabi-etal-2018-looking,thorne-etal-2018-fever}.
In other cases, rationales are collected for an existing dataset by different annotators \citep{zaidan-eisner-piatko-2008-machine,camburu-etal-2018-esnli,rajani-etal-2019-explain}.
As previously stated, such rationales may not be faithful.
Rationale length can vary from sub-sentence spans \citep{talmor-etal-2019-commonsenseqa} to multiple sentences \citep{lehman-etal-2019-inferring}.

Model rationales can be produced as an explicit training objective \citep{zaidan-eisner-piatko-2008-machine} or extracted as a post-hoc explanation.
Post-hoc methods typically assign token-level importance scores: attention weights are often used in attention-based models \citep{bahdanau2016neural}, gradient-based explanations are typical for differentiable models \citep{denil-etal-2015-extraction, shrikumar-etal-2017-learning}, and LIME is a model-agnostic method \citep{ribeiro-etal-2016-trust}.
We follow work using BERT's attention \citep{clark-etal-2019-bert,kovaleva-etal-2019-revealing} to extract rationales.

A model rationale is evaluated on faithfulness (if it is actually used to make a decision) and plausibility (if it is easily understood by humans).
Faithfulness can be measured by perturbing inputs marked as evidence and measuring change in outputs \citep{jain-wallace-2019-attention,serrano-smith-2019-attention}.
Plausibility can be measured through user studies, wherein users are given a model rationale and asked either to predict the model's decision \citep{kim-etal-2016-examples} or to rate rationale understandability \citep{nguyen-2018-comparing,ehsan-etal-2018-rationalization,ehsan-etal-2019-automated,strout-etal-2019-human}.
Rationale plausibility can also be measured by similarity to human rationales \citep{deyoung-etal-2020-eraser}, but this requires faithful human rationales.
We use similarity to evaluate rationale plausibility because we gather faithful human rationales from sentence editing annotations.

\section{Proposed Task}

We propose re-purposing the AESW classification task for measuring model interpretability.
We gather examples from AESW from which we can automatically extract faithful and sufficient human rationales, and then use said rationales to investigate factors in PLM rationale plausibility, specifically BERT \citep{devlin-etal-2019-bert} and its variants.
It is worth noting that our rationale dataset can be used for other interpretability topics such as training with rationales or evaluating rationale faithfulness; we will focus on plausibility considering the scope of this paper.

\subsection{Human Rationales}

Human rationales are substrings used as evidence for a decision \citep{zaidan-etal-2007-using}.
Faithful and sufficient (enough evidence to justify a decision) human rationales can be used as gold labels for evaluating model rationale plausibility.

The original AESW task is to predict if a sentence from a scientific paper needs editing.
\citeauthor{aesw} extract spans of a sentence before and after professional editing\footnote{
Native English speaking editors working at VTeX.
} as deleted (\token{<del>}) or inserted (\token{<ins>}) and provide 1.1M training, 140K validation and 140K testing examples.
Sentences without changes are assumed to not require editing.

We exploit the data format to automatically extract faithful and sufficient human rationales.
Delete text (text between \token{<del>} tags) is always a \textit{faithful} rationale (such text is a source of the error).
For edits with arbitrary \token{<del>} text alone and \token{<ins>} text, the \token{<del>} alone is not always a \textit{sufficient} rationale to justify \neededit{}. 
Consider a sentence where a verb is incorrectly conjugated and replaced with the correctly conjugated verb. 
The incorrectly conjugated verb is not sufficient to decide that the sentence needs editing; the surrounding context is required.
To find edits where \token{<del>} text \textit{is} always a sufficient rationale, we use two criteria:
\begin{enumerate}
    \item A spelling error is corrected. (\spelling{})
    \item Text is only deleted, not added. (\deleted{})
\end{enumerate}
Spelling errors are always a sufficient rationale to justify editing a sentence (see Figure \ref{fig:aesw-example}a).
In edits with no insertions, removing the \token{<del>} text leads to an error-free sentence, so the \token{<del>} text is sufficient explanation for editing (see Figure \ref{fig:aesw-example}b).
These criteria lead to both a simple, lexical task (\spelling) and a more challenging semantic task (\deleted) and constitute a wider range of challenges for future interpretability research. 
We extract faithful and sufficient human rationales for 1,321 \spelling{} edits and 6,741 \deleted{} edits from the validation set of AESW.\footnote{
    More details on extracting human rationales, as well as additional examples, can be found in Appendix \ref{app:human-rationale-details}
}

\subsection{Model Rationales}\label{sec:model-rationales}

Model rationales are substrings provided by a model as evidence for a decision.
Given a model, an example $x_i$ and a prediction $y_i$, we extract three model rationales.

First, we use attention maps \citep{xu-etal-2015-show,kovaleva-etal-2019-revealing} to rank word relevance.
We measure the total attention weight from BERT's final layer's initial \token{[CLS]} token to each token $t_j$ across $H$ attention heads.
Then we add those totals together for each token $t_j$ in a word $w$: 
\begin{equation}\label{eq:attention-weights}
\text{score}(w) = \sum_{t_j \in w}\sum_{h = 1}^{H} \text{Attn}_{h_j}\left( \text{[CLS]} \to t_j \right).
\end{equation}

Second, we use gradient-based saliency (specifically gradient$\times$input; \citealp{denil-etal-2015-extraction,shrikumar-etal-2017-learning}) to rank word relevance.
We calculate a saliency score for each token $t_j$ in $x_i$.
First, change in model output with respect to $t_j$'s input embedding $\nabla_{e(t_j)}f_{y_i}(x_i)$ captures the sensitivity to token $t$.
The dot product with $e(t_j)$ is then a scalar measure of each token's marginal impact on the model prediction \citep{han-etal-2020-explaining}.
Finally, we again compute a word-level score by summing over each token $t_j$ in word $w$:
\begin{equation}\label{eq:input-gradient}
\text{score}(w) = \sum_{t_j \in w}\nabla_{e(t_j)}f_{y_i}(x_i) \cdot e(t_j)
\end{equation}
In contrast to attention, \gradientxinput{} faithfully measures the marginal effect of each input token on the prediction \citep{bastings-filippova-2020-elephant}.
Finally, we extract a third set of rankings using gradient$\times$input's magnitude to rank words.

\subsection{Evaluation}

We evaluate extracted rationale plausibility using similarity to human rationales \citep{deyoung-etal-2020-eraser}.
We use the continuous word scores generated in the previous section to rank relevance, then use mean reciprocal rank as an evaluation metric.\footnote{
    We performed our analyses with additional, similar metrics; our main observations remain consistent independent of the choice of metric. Appendix \ref{app:model-rationale-details} contains additional details and complete results.
}

\section{Experiments}

To demonstrate the utility of the AESW task for interpretability research, we present four experiments, each with the goal of understanding factors in PLM rationale plausibility.


For our experiments, we add a linear layer and sigmoid activation function on top of the \token{[CLS]} token representation, fine-tune BERT-base end-to-end on the AESW training set using the original training objective (classify a sentence as \neededit{} or \noedit{}) and use validation loss to tune hyperparameters.
We do not add any interpretability or rationale-related objectives.
The classification F1 for the models on \spelling{} and \deleted{} edits is 82.0 and 74.0, respectively.\footnote{
Appendix \ref{app:training-details} 
contains more details on fine-tuning procedure and results.
}
We extract and evaluate model rationales on all \spelling{} and \deleted{} edits, even edits that a model does not correctly predict as \neededit{.}

\paragraph{Does pre-training procedure affect rationale plausibility?}
We are interested in how pre-training affects plausibility after fine-tuning (to the best of our knowledge, this is a previously unexplored topic).
We compare BERT and two variants, RoBERTa~\cite{liu2019roberta} and SciBERT~\cite{beltagy-etal-2019-scibert}. These models have the same architecture but differ in pre-training: RoBERTa is pre-trained for 5x longer with 10x more data than BERT, and SciBERT is pre-trained on a corpus of academic papers.
We extract model rationales using attention weights and evaluate their plausibility.

\begin{figure}[t]
  \includegraphics[width=\linewidth]{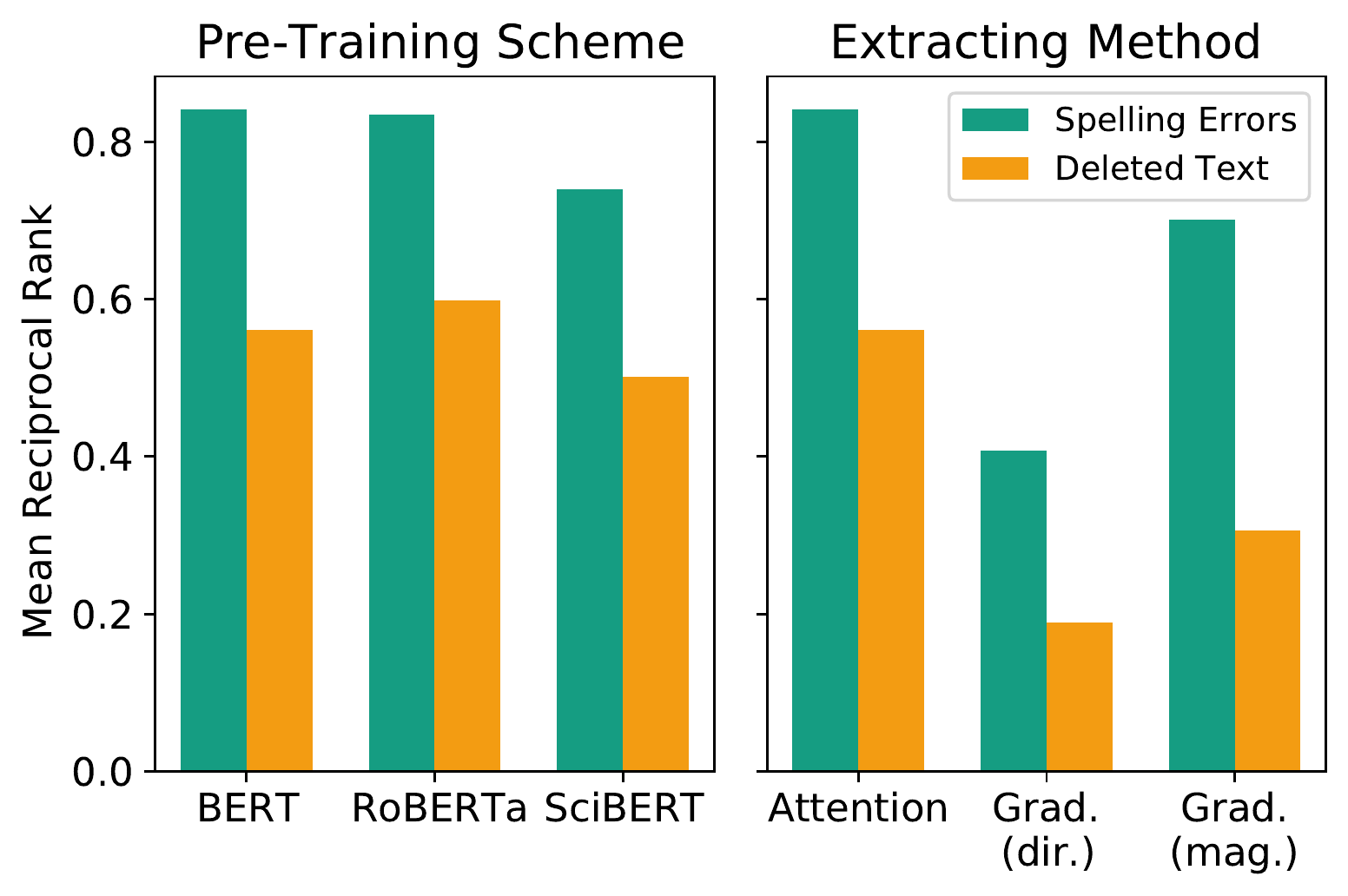}
  \caption{
    \textbf{Left}: BERT, RoBERTa and SciBERT's attention weight rationale plausibility.
    \textbf{Right}: BERT's attention weight, \gradientxinput{} and \gradientxinputmagnitude{} rationale plausibility (see Section \ref{sec:model-rationales}).
  }
  \label{fig:big-bar-chart}
  \vspace{-10pt}
\end{figure}

As seen in Figure \ref{fig:big-bar-chart}, RoBERTa and BERT generate nearly equally plausible rationales despite differences in pre-training corpus size.
We hypothesize that SciBERT generates less plausible rationales because it encodes \neededit{} representations in earlier layers (rather than in the final layer), then attends to \token{[SEP]} as a no-op in later layers, as proposed in \citet{clark-etal-2019-bert} and \citet{kobayashi-etal-2020-attention} and further confirmed in the subsequent experiments. 

\paragraph{Do attention weights or input gradients produce better rationales?}
In contrast to attention weights, \gradientxinput{} scores are naturally faithful with respect to individual feature importance \citep{jain-wallace-2019-attention,bastings-filippova-2020-elephant}.
However, attention weights can represent word relevance \textit{in context}, potentially leading to more plausible rationales. We extract and evaluate rationales using attention weights and the two \gradientxinput{} methods described in Section \ref{sec:model-rationales}.

Figure \ref{fig:big-bar-chart} shows that attention-based rationales are more plausible and that the difference is more pronounced on \deleted{} edits.
Using \gradientxinputmagnitude{} (right-most) also shows improvements over directional \gradientxinput{} (middle-right), in contrast to \citet{han-etal-2020-explaining}.

\paragraph{Are plausibility and confidence correlated?}
We are curious if BERT is more confident in its classification decisions when it attends to the original evidence used by editors.
We quantify BERT's classification confidence by treating the sigmoid activation function's value as the probability of a \neededit{} decision.
We only consider examples that BERT correctly classifies as \neededit{} and calculate BERT's mean confidence.
We find that when human rationales are BERT's most attended-to words, it is 11.7\% and 11.9\% more confident in its predictions for \spelling{} and \deleted{} edits, respectively.

One possible explanation for such a correlation is that easy-to-classify examples also have easy-to-identify rationales.
Because agreement with human rationales is not the training objective, however, we believe this correlation suggests that BERT learns to classify edits similarly to humans.

\begin{figure}[t]
  \includegraphics[width=\linewidth]{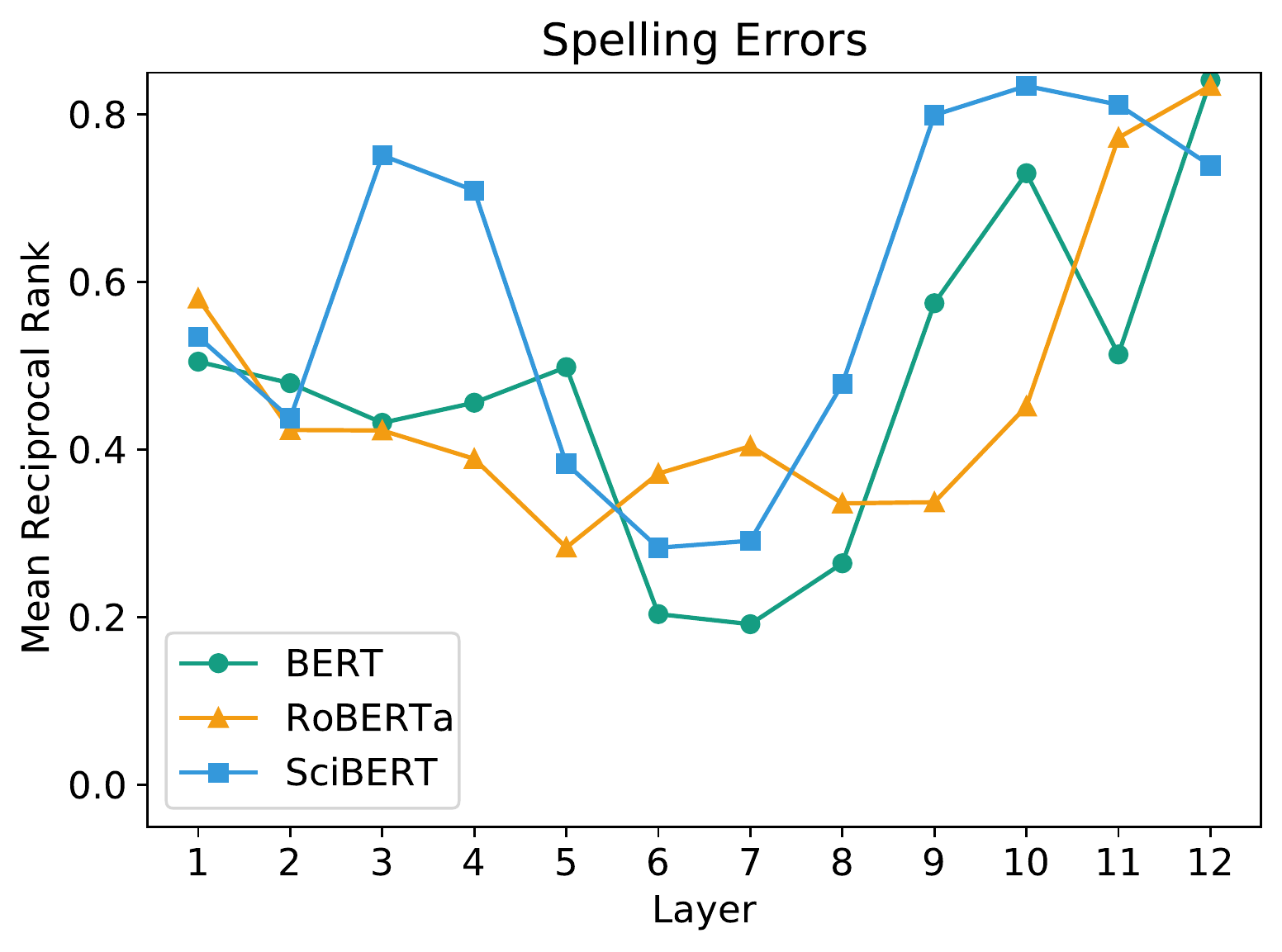}
  \includegraphics[width=\linewidth]{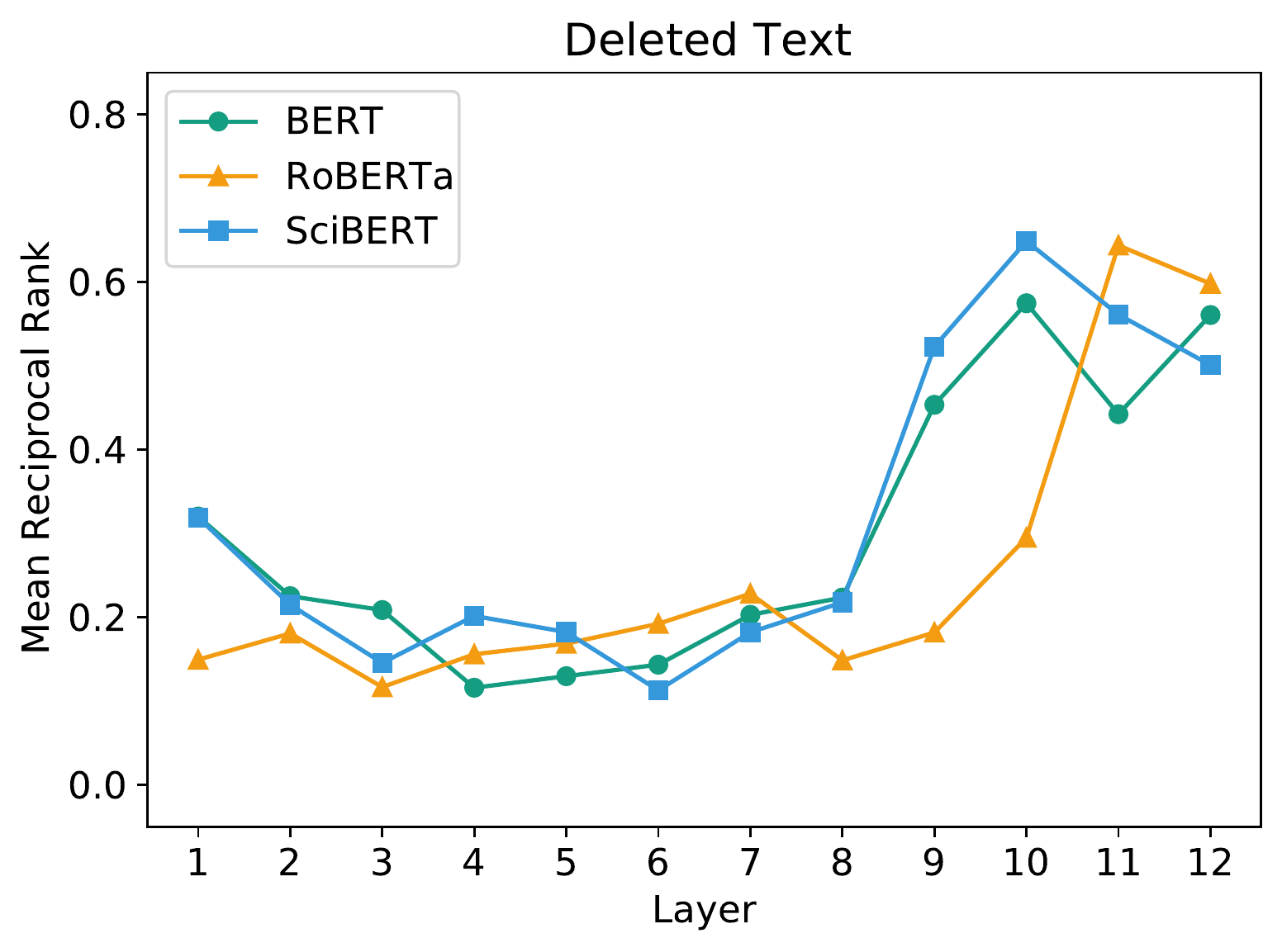}
  \caption{
    Mean reciprocal rank for each layer (using the \textit{mean} strategy) for each model for \spelling{} and \deleted{} edits.\footnotemark[4]
  }
  \label{fig:layers}
  \vspace{-10pt}
\end{figure}

\paragraph{How does transformer layer affect plausibility?}
It is widely agreed that BERT's early layers encode more lexical and phrasal information than other layers \citep{jawahar-etal-2019-bert, rogers-etal-2020-primer}.
We hypothesize that rationales extracted from early layers will be more plausible for \spelling{} than \deleted{} edits because it is a more lexical task.
We extract rationales from each layer's attention weights and measure their plausibility in Figure \ref{fig:layers}.

\footnotetext[4]{
BERT shows a decline in plausibility at layer 11 in both edit types because it attends heavily to periods. 
}

The results confirm the hypothesis and show that early layers in BERT models are indeed more lexically-oriented. We also find that SciBERT strongly attends to spelling errors in earlier layers.
We believe it is because that, pre-trained on well-formed academic text, SciBERT is not as well exposed to spelling errors as BERT and RoBERTa so it learns to attend to spelling errors in earlier layers during fine-tuning.


\section{Conclusion}

We re-purpose the AESW task to gather thousands of inherently faithful human rationales and investigate an array of questions regarding PLM interpretability.
We find, among other new insights, that attention weights correlate well with human rationales and produce more plausible rationales than input gradients, which is different from existing understanding. Furthermore, we find that BERT is more confident in its predictions when it attends to the same words that a human did, supporting the idea that while attention is not inherently faithful, attention-based models might rely on the same information as humans when making a prediction.

Future work might expand the subset of examples for which human rationales can be automatically extracted, include human rationales during training or evaluate faithful-by-design model rationales on this dataset.

\section{Acknowledgements}

The authors would like to extend their thanks to the anonymous reviewers for their insightful feedback.

\bibliography{anthology,custom,acl_anthology}
\bibliographystyle{acl_natbib}

\appendix

\begin{appendix}
\appendix
\clearpage

\section{AESW Details}

The original AESW task is to classify a sentence from an academic paper as \neededit{} or \noedit{}.
There is no special markup attached to the sentence when it is given as input to a model.
Task participants receive the training and validation sets with \token{<del>} and \token{<ins>} tokens for model development, and \citeauthor{aesw} provided an automatic test set evaluation through an online portal.
\citeauthor{aesw} collected these sentences from 9,919 journal articles published by Springer Publishing Company and edited at VTeX. 
Sentences before and after editing were automatically aligned using a modified diff algorithm.\footnote{More details can be found in \cite{aesw}}
After the challenge, \citeauthor{aesw} released the dataset in its entirety (all three datasets with all tokens/spans included).

\section{Training Details} \label{app:training-details}

The AESW task uses scientific papers written in LaTeX, which contains markup characters that impact sentence meaning.
The original authors (\citeauthor{aesw}) replace these characters with special tokens, as seen in Table \ref{tab:special-tokens}.
We add these four special tokens (\MATH, \_MATHDISP\_, \_CITE\_ and \_REF\_) to the model vocabulary, fine-tuning the word representations during training.

\begin{table}[h]
  \centering
  \begin{tabular}{l l} \hline
    \textbf{LaTeX Example} & \textbf{Special Token} \\ \hline
    \texttt{\$\textbackslash beta\_\{2\}\$} & \MATH \\
    \texttt{\$\$2 + 3\$\$} & \_MATHDISP\_ \\
    \texttt{\textbackslash cite\{google2018\}} & \_CITE\_ \\
    \texttt{\textbackslash ref\{tab:results\}} & \_REF\_ \\ \hline
  \end{tabular}
  \caption{Special tokens found in the original AESW data that should not be split further into bytes/tokens.}
  \label{tab:special-tokens}
\end{table}

We train all models for a maximum of 30 epochs with a patience of 5 on a single Tesla P100 GPU.

All models (BERT\footnote{\url{https://huggingface.co/transformers/v3.0.2/model\_doc/bert.html\#bertforsequenceclassification}}, SciBERT\footnote{\url{https://github.com/allenai/scibert\#pytorch-huggingface-models}}, RoBERTA\footnote{\url{https://huggingface.co/transformers/v3.0.2/model\_doc/roberta.html\#robertaforsequenceclassification}}) are based on their HuggingFace \citep{Wolf2019HuggingFacesTS} implementations.

We list all the key hyperparameters and tuning bounds for reproducibility in Table \ref{tab:hyperparameters}.
Our final results for all three BERT-based models along with the top three models from \cite{aesw} can be found in Table \ref{tab:aesw-results}.
Additionally, we will release code and instructions for reproducing our results.

\begin{table}[ht]
  \centering
  \small
  \begin{tabular}{lcc}
    \toprule
    \textbf{Model} & \textbf{Dev F1} & \textbf{Test F1} \\ 
    \midrule
    \textbf{CNN+LSTM} & -- & 0.628 \\
    \textbf{CNN} & -- & 0.611 \\ 
    \textbf{SVM} & -- & 0.555 \\
    \midrule
    \textbf{BERT}$_{\textbf{base}}$      & 0.654 & 0.666 \\
    \textbf{RoBERTa}$_{\textbf{base}}$   & \textbf{0.661} & \textbf{0.670} \\
    \textbf{SciBERT}$_{\textbf{scivocab}}$   & 0.658 & 0.668 \\
    \bottomrule
  \end{tabular}
  \caption{Performance on the original AESW sentence classification task. Dev set results are not available for models reported in \citet{aesw}.}
  \label{tab:aesw-results}
  \vspace{-10pt}
\end{table}

\begin{table*}
  \centering
  \begin{tabular}{l l l l} \hline
    \textbf{Model} & \textbf{Hyperparameters} & \textbf{Hyperparameter bounds} \\ \hline
    \textbf{BERT$_{\text{base}}$} & \makecell[l]{
      learning rate: $1 \times 10^{-6}$ \\ 
      batch size: 32 \\ 
      model: bert-base-uncased \\ 
      vocab size: 30526 (normally 30522)
    } &  \makecell[l]{
      learning rate: ($2 \times 10^{-7}$, \\ 
      $1 \times 10^{-6}$, $2 \times 10^{-5}$, \\ 
      $1 \times 10^{-4}$)
    } \\
    \textbf{RoBERTa$_{\text{base}}$} & \makecell[l]{
      learning rate: $1 \times 10^{-6}$ \\ 
      batch size: 32 \\ 
      model: roberta-base \\ 
      vocab size: 50269 (normally 50265)
    } & \makecell[l]{
      learning rate: ($1 \times 10^{-6}$)
    } \\ 
    \textbf{SciBERT} & 
    \makecell[l]{
      learning rate: $1 \times 10^{-6}$ \\ 
      batch size: 32 \\ 
      model: allenai/scibert\_scivocab\_uncased \\ 
      vocab size: 31094 (normally 31090)
    } & \makecell[l]{
      learning rate: ($1 \times 10^{-6}$)
    } \\ \hline
  \end{tabular}
  \caption{Hyperparameter options for each model. Note that each model had 4 special tokens added to the vocabulary. BERT was fine-tuned first. Because of compute limitations, RobERTa and SciBERT were both fine-tuned using the same hyperparameters as the optimal BERT configuration (learning rate of ($1 \times 10^{-6}$)).}
  \label{tab:hyperparameters}
\end{table*}

\section{Gathering Human Rationales} \label{app:human-rationale-details}

To find \spelling{} edits, we look for sentences with a deleted, misspelled word followed by an inserted, correctly spelled word. 
The first two examples in Figure \ref{fig:detailed-aesw-examples} are examples of \spelling{} edits, while the third is not.

To find \deleted{} edits, we look for sentences where text is removed but not added. 
The fourth and fifth examples in Figure \ref{fig:detailed-aesw-examples} are examples of \deleted{} edits.

\begin{figure*}[t]
  \centering
  \begin{tabularx}{\linewidth}{X}
    \toprule
    \texttt{\nop{<sentence sid="8447.0">}The algorithm <del>descripted</del><ins>described</ins> in the previous sections has several advantages.\nop{</sentence>}} \\
    A \spelling{} edit; ``descripted'' is a spelling error and is corrected. \\ 
    \midrule
    \texttt{\nop{<sentence sid="256.4">}For each energy point, the thereby obtained cross-section values and errors from different experiments have been further averaged according to the <del>weigthed</del><ins>weighted</ins> average method used by the Particle Data Group \_CITE\_, including error rescaling by \MATH{} in case of large discrepancy.\nop{</sentence>}} \\
    A \spelling{} edit; ``weigthed'' is a spelling error and is corrected. \\
    \midrule
    \texttt{\nop{<sentence sid="49.2">}And the short notations for the denominators are \_MATHDISP\_, Furthermore, the following relations are useful to <del>short</del><ins>shortcut</ins> the expressions: \_MATHDISP\_.\nop{</sentence>}} \\
    \textbf{Not} a spelling error edit; ``short'' is incorrect in this context, but it is not a spelling error. \\
    \midrule
    \midrule
    \texttt{\nop{<sentence sid="662.3">}However, we <del>must note that we </del>still have no means of deciding which documents out of \MATH{} deserve to be in \MATH{} and \MATH{}, respectively.\nop{</sentence>}} \\ 
    A \deleted{} edit; text is only removed, while no text is inserted. \\
    \midrule
    \texttt{\nop{<sentence sid="519.5">}Let \MATH{} be the conjugate Holder<del>'s</del> function of a Holder<del>'s</del> function \MATH{.}\nop{</sentence>}} \\
    A \deleted{} edit; text is only removed, while no text is inserted. \\ 
    \bottomrule
  \end{tabularx}
  \caption{Additional examples of the two types of edits extracted from the original AESW dataset.}
  \label{fig:detailed-aesw-examples}
\end{figure*}

\section{Evaluating Model Rationales} \label{app:model-rationale-details}

Although the main text presents our analyses using mean reciprocal rank, we performed our analyses with multiple metrics.
Here we provide specific definitions for our metrics and our complete results for all models with every metric.

\paragraph{Mean reciprocal rank}


Because human rationales can be made up of multiple words, we need to modify mean reciprocal ranking.
For a single sentence, a model's ordered ranking of words $M$ and an unordered human rationale $H$:
\begin{enumerate}
    \item Find the top ranked word $w$ in $M$ from $H$ and record the rank.
    \item Remove $w$ from $M$ and $H$.
    \item Repeat until $H$ is empty.
    \item Use the reciprocal of the mean rank.
\end{enumerate}
If we did not remove words from $M$, a perfect score for a rationale with multiple words would be impossible.
Consider the sentence ``The boy \sout{ate the} \textbf{found his} ball.''
If the rationale ranked `ate' and `the' as most important, the mean rationale rank would be $(1 + 2) / 2 = 1.5$ and the rationale's reciprocal rank would be $\sfrac{1}{1.5} = 0.66$, despite a perfect rationale.

We take the mean reciprocal ranking across all examples to evaluate rationales.

\paragraph{Mean area under precision-recall curve}

Using the model relevance scores for words in a sentence, we adjust the threshold for classifying a word as part of the rationale, calculate a precision-recall curve and measure the area underneath.
We take the mean AUPRC across all examples.

\paragraph{Mean top 1 match}

We score a rationale as 1 if the rationale's top ranked word is the human rationale and 0 otherwise.
This means that all multiple-word rationales are automatically not a match and scored as 0.
We take the mean of these scores across all examples.

\medskip
Our main observations are consistent across metrics: Table \ref{tab:spelling-errors-results} contains all results for models on \spelling{} edits and Table \ref{tab:deleted-text-results} contains all results for models on \deleted{} edits.

\begin{table*}
\centering
    \begin{tabular}{l r l l r r r}
        \toprule
        Model & F1 & Method & Classification & Mean Recip. & AUPRC & Top 1 Match \\ 
        \midrule
BERT & 82.1 & Attention & All & 0.841 & 0.115 & 0.756 \\
BERT & 100.0 & Attention & Correct & 0.895 & 0.081 & 0.824 \\
BERT & 0.0 & Attention & Wrong & 0.717 & 0.194 & 0.602 \\
BERT & 82.1 & \gradientxinput{} & All & 0.407 & 0.332 & 0.325 \\
BERT & 100.0 & \gradientxinput{} & Correct & 0.401 & 0.335 & 0.322 \\
BERT & 0.0 & \gradientxinput{} & Wrong & 0.422 & 0.325 & 0.331 \\
BERT & 82.1 & \gradientxinputmagnitude{} & All & 0.701 & 0.201 & 0.579 \\
BERT & 100.0 & \gradientxinputmagnitude{} & Correct & 0.730 & 0.184 & 0.615 \\
BERT & 0.0 & \gradientxinputmagnitude{} & Wrong & 0.635 & 0.241 & 0.498 \\
\midrule
RoBERTa & 81.4 & Attention & All & 0.834 & 0.118 & 0.739 \\
RoBERTa & 100.0 & Attention & Correct & 0.886 & 0.085 & 0.811 \\
RoBERTa & 0.0 & Attention & Wrong & 0.721 & 0.192 & 0.581 \\
RoBERTa & 81.4 & \gradientxinput{} & All & 0.310 & 0.380 & 0.231 \\
RoBERTa & 100.0 & \gradientxinput{} & Correct & 0.303 & 0.383 & 0.229 \\
RoBERTa & 0.0 & \gradientxinput{} & Wrong & 0.324 & 0.375 & 0.236 \\
RoBERTa & 81.4 & \gradientxinputmagnitude{} & All & 0.675 & 0.219 & 0.546 \\
RoBERTa & 100.0 & \gradientxinputmagnitude{} & Correct & 0.707 & 0.201 & 0.583 \\
RoBERTa & 0.0 & \gradientxinputmagnitude{} & Wrong & 0.605 & 0.259 & 0.463 \\
\midrule
SciBERT & 86.8 & Attention & All & 0.739 & 0.203 & 0.577 \\
SciBERT & 100.0 & Attention & Correct & 0.830 & 0.138 & 0.707 \\
SciBERT & 0.0 & Attention & Wrong & 0.442 & 0.415 & 0.152 \\
SciBERT & 86.8 & \gradientxinput{} & All & 0.475 & 0.296 & 0.399 \\
SciBERT & 100.0 & \gradientxinput{} & Correct & 0.467 & 0.296 & 0.401 \\
SciBERT & 0.0 & \gradientxinput{} & Wrong & 0.501 & 0.295 & 0.392 \\
SciBERT & 86.8 & \gradientxinputmagnitude{} & All & 0.670 & 0.214 & 0.557 \\
SciBERT & 100.0 & \gradientxinputmagnitude{} & Correct & 0.698 & 0.197 & 0.591 \\
SciBERT & 0.0 & \gradientxinputmagnitude{} & Wrong & 0.575 & 0.268 & 0.447 \\
        \bottomrule
    \end{tabular}
    \caption{Results for all three models for \spelling{} edits across all methods of selecting rationales (attention, \gradientxinput{} and \gradientxinputmagnitude{}) and all metrics to evaluate rationales (mean reciprocal ranking, AUPRC and mean top 1 match).}
  \label{tab:spelling-errors-results}
\end{table*}

\begin{table*}
\centering
    \begin{tabular}{l r l l r r r}
        \toprule
        Model & F1 & Method & Classification & Mean Recip. & AUPRC & Top 1 Match \\ 
        \midrule
BERT & 74.0 & Attention & All & 0.561 & 0.272 & 0.411 \\
BERT & 100.0 & Attention & Correct & 0.680 & 0.211 & 0.528 \\
BERT & 0.0 & Attention & Wrong & 0.391 & 0.359 & 0.244 \\
BERT & 74.0 & \gradientxinput{} & All & 0.189 & 0.447 & 0.074 \\
BERT & 100.0 & \gradientxinput{} & Correct & 0.195 & 0.442 & 0.089 \\
BERT & 0.0 & \gradientxinput{} & Wrong & 0.181 & 0.454 & 0.054 \\
BERT & 74.0 & \gradientxinputmagnitude{} & All & 0.306 & 0.416 & 0.130 \\
BERT & 100.0 & \gradientxinputmagnitude{} & Correct & 0.354 & 0.399 & 0.166 \\
BERT & 0.0 & \gradientxinputmagnitude{} & Wrong & 0.238 & 0.440 & 0.078 \\
\midrule
RoBERTa & 74.8 & Attention & All & 0.598 & 0.263 & 0.426 \\
RoBERTa & 100.0 & Attention & Correct & 0.662 & 0.230 & 0.487 \\
RoBERTa & 0.0 & Attention & Wrong & 0.503 & 0.312 & 0.335 \\
RoBERTa & 74.8 & \gradientxinput{} & All & 0.168 & 0.456 & 0.058 \\
RoBERTa & 100.0 & \gradientxinput{} & Correct & 0.156 & 0.458 & 0.054 \\
RoBERTa & 0.0 & \gradientxinput{} & Wrong & 0.187 & 0.452 & 0.063 \\
RoBERTa & 74.8 & \gradientxinputmagnitude{} & All & 0.321 & 0.404 & 0.154 \\
RoBERTa & 100.0 & \gradientxinputmagnitude{} & Correct & 0.367 & 0.384 & 0.194 \\
RoBERTa & 0.0 & \gradientxinputmagnitude{} & Wrong & 0.252 & 0.434 & 0.094 \\
\midrule
SciBERT & 73.9 & Attention & All & 0.501 & 0.327 & 0.304 \\
SciBERT & 100.0 & Attention & Correct & 0.605 & 0.270 & 0.414 \\
SciBERT & 0.0 & Attention & Wrong & 0.353 & 0.406 & 0.148 \\
SciBERT & 73.9 & \gradientxinput{} & All & 0.226 & 0.426 & 0.111 \\
SciBERT & 100.0 & \gradientxinput{} & Correct & 0.260 & 0.410 & 0.145 \\
SciBERT & 0.0 & \gradientxinput{} & Wrong & 0.178 & 0.450 & 0.062 \\
SciBERT & 73.9 & \gradientxinputmagnitude{} & All & 0.322 & 0.400 & 0.162 \\
SciBERT & 100.0 & \gradientxinputmagnitude{} & Correct & 0.356 & 0.387 & 0.189 \\
SciBERT & 0.0 & \gradientxinputmagnitude{} & Wrong & 0.275 & 0.418 & 0.123 \\
        \bottomrule
    \end{tabular}
    \caption{Results for all three models for \deleted{} edits across all methods of selecting rationales (attention, \gradientxinput{} and \gradientxinputmagnitude{}) and all metrics to evaluate rationales (mean reciprocal ranking, AUPRC and mean top 1 match).}
  \label{tab:deleted-text-results}
\end{table*}

\end{appendix}

\end{document}